\documentclass{article} 
\usepackage{style,times}
\usepackage{booktabs}
\usepackage{multirow}
\usepackage{mathtools}

\usepackage{amsmath,amsfonts,bm}

\def\eqref#1{equation~\ref{#1}}

\def\1{\bm{1}}

\DeclareMathAlphabet{\mathsfit}{\encodingdefault}{\sfdefault}{m}{sl}
\SetMathAlphabet{\mathsfit}{bold}{\encodingdefault}{\sfdefault}{bx}{n}

\usepackage{hyperref}
\usepackage{url}

\usepackage{xcolor}

\title{Learning Generation Orders for Masked Discrete Diffusion Models via Variational Inference}

\author{
  David Fox \thanks{Corresponding author.}\\
  University of Bristol\\
  \texttt{david.k.fox@bristol.ac.uk}
  \And
  \hspace{-2.5in}Sam Bowyer \\
  \hspace{-2.5in}University of Bristol\\
  \hspace{-2.5in}\texttt{sam.bowyer@bristol.ac.uk}
  \AND 
  Song Liu \\
  University of Bristol\\
  \texttt{song.liu@bristol.ac.uk}
  \And 
  Laurence Aitchison \\
  Mistral AI\\
  \texttt{laurence.aitchison@gmail.com}
  \AND 
  Raul Santos-Rodriguez \\
  University of Bristol\\
  \texttt{enrsr@bristol.ac.uk}
  \And
  \hspace{1in}Mengyue Yang \\
  \hspace{1in}University of Bristol\\
  \hspace{1in}\texttt{mengyue.yang@bristol.ac.uk}
}

\iclrfinalcopy 
\begin{document}

\maketitle

\begin{abstract}
Masked discrete diffusion models (MDMs) are a promising new approach to generative modelling, offering the ability for parallel token generation and therefore greater efficiency than autoregressive counterparts. However, achieving an optimal balance between parallel generation and sample quality remains an open problem. Current approaches primarily address this issue through fixed, heuristic parallel sampling methods. There exist some recent learning based approaches to this problem, but its formulation from the perspective of variational inference remains underexplored. In this work, we propose a variational inference framework for learning parallel generation orders for MDMs. As part of our method, we propose a parameterisation for the approximate posterior of generation orders which facilitates parallelism and efficient sampling during training. Using this method, we conduct preliminary experiments on the GSM8K dataset, where our method performs competitively against heuristic sampling strategies in the regime of highly parallel generation. For example, our method achieves 33.1\% accuracy with an average of only only 4 generation steps, compared to 23.7-29.0\% accuracy achieved by standard competitor methods in the same number of steps. We believe further experiments and analysis of the method will yield valuable insights into the problem of parallel generation with MDMs.
\end{abstract}

\section{Introduction}
Discrete Diffusion Models (DDMs) \cite{austin2021structured} have recently emerged as a promising alternative to autoregressive generative models for various tasks such as text \cite{song2025seed}, code \cite{khanna2025mercury} and biological sequence generation \cite{vignac2022digress}, offering the ability to parallelise token generation for increased efficiency \cite{hu2025accelerating} and utilise bi-directional context during generation which has shown to be beneficial in certain tasks where a strict left-to-right dependency is not present in the data \cite{nie2025large}. Masked Diffusion Models (MDMs) \cite{austin2021structured} in particular have established themselves as the most performant class of DDMs, due to their comparatively efficient paramaterisation and conceptually simple set-up \cite{shi2024simplified, sahoo2024simple}, as well as their similarity with Autoregressive Models (ARMs), making it possible to initialise training from strong autoregressive baselines \cite{ye2025dream}. 

Despite the conceptual benefits and recent empirical progress of DDMs \cite{nieScalingMaskedDiffusion2025}, a consistent barrier to realising their full potential has been optimally balancing efficiency and sample quality by choosing generation orders which use parallelization without violating statistical dependence among token positions \cite{kang2025parallelbench}. 

Current approaches attempt to overcome this issue by using heuristic sampling strategies, or by augmenting the generative model with a learned component which chooses token positions to unmask at each step. Heuristic approaches typically use model predicted logits to select tokens based on criteria such as top-k or top probability margin \cite{kimTrainWorstPlan2025}. Learned approaches usually augment the generative model to include a component for selecting token indices which is then trained with a separate loss function \cite{zhangzhi2025path}, or via reinforcement learning by formulating the generative model as a Markov Decision Process with a policy that chooses token positions to unmask \cite{hong2025improving}. 

While these methods have made progress toward achieving the optimal trade-off between generation efficiency and model output quality, we believe there is room for improvement. While heuristic approaches provide good performance at minimal cost, they are potentially too rigid and overly reliant on logit based model confidence estimates, which my be poorly calibrated for this purpose when trained using only binary cross-entropy. Of the learned approaches, the formulation of learned parallel generation orders within the variational inference framework remains underexplored.

Hence, the goal of the present work is to investigate a variational inference framework for training an MDM which explicitly factorises the model into components for choosing which token positions to unmask, and which token value to sample given a position. We believe this formulation offers the following benefits which may lead to a method that allows training of an improved unmasking position model which scales to large datasets. Our main contributions are as follows: 

\begin{itemize}
    \item We present a probabilistic formulation of a discrete diffusion generative model using variational inference, which explicitly factorises the model into components for choosing which token positions to unmask, and which token value to sample given a position. 
    \item We derive the associated ELBO objective, which leverages this model structure to decrease variance of the objective function through Rao-Blackwellisation. 
    \item We investigate the use of a paramaterised family of distributions for the approximate posterior generation order which is designed to allow efficient, low variance training.
\end{itemize}

The remainder of this document is structured as follows: In section \ref{sec:background} we discuss necessary background information on masked DDMs, the issue of accurately sampling in parallel, and some current approaches to overcoming this issue. In section \ref{sec:method} we discuss our method, which is motivated by these concerns, and a plan for experiments to test it, which are reported in section \ref{sec:experiments} and discussed in section \ref{sec:discussion}. 

\section{Background}\label{sec:background}
\subsection{Masked Discrete Diffusion Models}
To construct an MDM \cite{shi2024simplified} for a single dimension, we define a forward Markov process $\{x_t\}_{t \in \left[0,1\right]}$, taking values in a discrete set $x_t \in \{1, \dots, m-1, m\}$, where $m$ represents a special 'mask' token. Letting $\mathbf{1}$ denote a vector of ones and $\textbf{x}_s$ the one hot encoding of $x_s$, the forward process transition probabilities between times $s < t$ are defined as 

\begin{equation}
  \begin{dcases}
    q_{t|s}(x_t|x_s) = \text{Cat}(x_t; \bar{Q}(s,t)^T \textbf{x}_s), \\
    \bar{Q}(s,t) = \frac{\alpha_t}{\alpha_s} I + \left(1-\frac{\alpha_t}{\alpha_s}\right)\mathbf{1}\textbf{e}_m^T,
  \end{dcases}
\end{equation}

and the initial distribution at $t=0$ is the data distribution. The generative model is then defined as an approximation of the time reversal of the forward process. Paramaterising this requires the reverse time transition probability conditional on a clean data sample $x_0$,

\begin{equation}
  \begin{dcases}
     q_{s|t,0}(x_s|x_t, x_0) = \text{Cat}(x_s; \bar{R}(t,s)^T \textbf{x}_t), \\
    \bar{R}^{x_0}(t,s) = I + \frac{\alpha_s - \alpha_t}{1 - \alpha_t} \textbf{e}_m\left(\textbf{x}_0 - \textbf{e}_m\right)^T,
  \end{dcases}
\end{equation}

which is combined with a learned approximation $\mu_{\theta}$ of $q_{0|t}$ to give the reverse transition probability,

\begin{equation}
    p_{\theta}\left(x_s \middle| x_t\right) = q_{s|t,0}\left(x_s \middle| x_t, \mu_{\theta}(x_t, t)\right).
\end{equation} 

To extend to the multi-dimensional case $x_t$ = $x_t^{1}...x_{t}^{N}$ without sacrificing computational tractability, we use the following conditional independence assumptions, 

\begin{equation}
  \begin{dcases}
        q_{t|s}(x_t | x_s) = \prod_{n=1}^{N} q_{t|s}\left(x_t^{n} \middle| x_s^{n}\right), \\
    q_{s|t,0}\left(x_s \middle| x_t, x_0\right) = \prod_{n=1}^{N} q_{{s|t,0}}\left(x_s^{n} \middle| x_t^{n}, x_0^{n}\right), \\
        p_{\theta}\left(x_s \middle| x_t\right) = \prod_{n=1}^{N} q_{s|t,0}\left(x_s^{n} \middle| x_t^{n}, \mu_{\theta}^{n}(x_t, t)\right),
  \end{dcases}
\end{equation}

where $\mu_{\theta}^{n}(x_t, t)$ is trained to approximate $q_{0|t}(x_{0}^{n} | x_{t})$. The model weights $\theta$ are trained by minimising the ELBO for the model likelihood, using the forward model $Q$ as a fixed variational posterior. 

We can generate samples from the model via ancestral sampling using $p_\theta$ at discrete time points. Specifically, we discretise $[0, 1]$ uniformly into $T$ points, $\{t/T \}_{t=0}^{T}$, initialise $\textbf{x}_1 = m...m$, and iteratively sample the $\textbf{x}_{t}$'s according to $p_{\theta}\left(x_s \middle| x_t\right)$ \footnote{We use the shorthand $x_{t} = x_{t/T}$}

\begin{align*}
    \textbf{x}_{t} &\sim  \prod_{n=1}^{N} q\left(x^n_{t} \middle| x_{t+1}^{n}, \mu_{\theta}^{n}(\textbf{x}_{t+1}, t)\right), \quad t = T-1, ..., 0 \\
    \textbf{x}_T &= (m...m)
\end{align*}

\subsection{Reparamaterised Discrete Diffusion}
As noted in previous work \cite{zheng2023reparameterized}, this time discretised model can be reparamaterised to explicitly include i.i.d binary token selection variables $r_t^{n}$, in a way that preserves the marginals of $\textbf{x}_{0:T}$, as follows 

\begin{equation}\label{basic_generative_model}
    P_{\theta}(\mathbf{x}_{0:T}, \mathbf{r}_{0:T-1}) = \delta_{(m...m)}( \textbf{x}_T)\prod_{t=0}^{T-1}\left[ \prod_{n=1}^{N} P_{\theta}(x_t^{n} | r_t^{n}, \mathbf{x}_{t+1}) \right] \left[ \prod_{n=1}^{N} P(r_t^{n}|\textbf{x}_{t}) \right]
\end{equation}

where 

\begin{align*}
     P(r_t^{n}|\textbf{x}_{t}) &= \text{Bern}\left(\frac{\alpha_{t} - \alpha_{t+1}}{1 - \alpha_{t+1}} \textbf{1}_{x_{t+1}^{n}=m}\right), \\
     P_{\theta}(x_t^{n} | r_t^{n}, \mathbf{x}_{t+1}) &= \delta_{x_{t+1}^{n}}(1 - r_t^{n}) + \mu_{\theta}^{n}(\textbf{x}_{t+1}, t) r_t^{n}.
\end{align*}

This paramaterisation is useful as the explicit separation between components allows us to better understand the importance of the method we choose to select token positions to unmask. Notably, the authors of \cite{ben-hamuAcceleratedSamplingMasked2025} use this form to show that the error between the ground truth and generative model distributions can be decomposed into a term for error in the denoising network and the mutual information between simultaneously generated token position distributions. 

Furthermore, this decomposition provides a template for more generally paramaterising a learnable token selection component. Indeed, given that DDMs are already trained via ELBO optimisation by considering $\textbf{x}_{1:T}$ as latent variable models, a natural extension is to follow the same procedure, but considering $\mathbf{r}_{0:T-1}$ as additional latent variables. Similar approaches have been applied successfully to learning generation orders for any-order autoregressive models \cite{wang2025learning}.

\section{Learning Generation Orders for Discrete Diffusion Models}\label{sec:method}
We formulate training an MDM with a learned token selection component as variational inference of a latent variable model through ELBO optimisation.

\subsection{Generative Model \& Approximate Posterior}
The generative model has a similar structure to \eqref{basic_generative_model}, 

\begin{equation}\label{generative_model}
    P_{\theta}(\mathbf{x}_{0:T}, \mathbf{r}_{0:T-1}) = \delta_{(m...m)}( \textbf{x}_T)\prod_{t=0}^{T-1}\left[ \prod_{n=1}^{N} P_{\theta}(x_t^{n} | r_t^{n}, \mathbf{x}_{t+1}) \right] \left[ \prod_{n=1}^{N} P_{\psi}(r_t^{n}|\textbf{x}_{t}) \right],
\end{equation}

but with a learned distribution for the unmasking variables $r^n_t$ incorporated,

\begin{equation}
 P_{\Psi}(r_t^{n} | \mathbf{x}_{t+1}) = \text{Bern}\left(p_{\psi}^{t, n}(\mathbf{x}_{t+1}) \textbf{1}_{x_{t+1}^{n}=m}\right).
\end{equation}

The approximate posterior used for variational inference has a similar form to that of the generative model, 

\begin{equation}\label{eq:q_VI_full}
    Q_{\phi}(\mathbf{x}_{0:T}, \mathbf{r}_{0:T}) = \delta_{(m...m)}( \textbf{x}_T)\prod_{t=1}^{T-1}\left[ \prod_{n=1}^{N} Q(x_t^{n} | r_t^{n}, x^{n}_{t+1}, x_0^{n}) \right] \left[ \prod_{n=1}^{N} Q_{\phi}(r^{n}_t | \textbf{x}_{t+1}, \textbf{x}_0) \right],
\end{equation}

where 

\begin{align}
    Q(x_t^{n} | r_t^{n}, x_{t+1}^{n}, x_0^{n}) &= \delta_{x_{t+1}^{n}}(1 - r_t^{n}) + \delta_{x_0^{n}} {r_t^{n}}, \\
    \quad Q_{\phi}(r_t^{n} | \mathbf{x}_{t+1}, \mathbf{x}_0) &= \text{Bern}\left(q_{\phi}^{t, n}(\mathbf{x}_{t+1}, \mathbf{x}_0) \textbf{1}_{x_{t+1}^{n}=m}\right).
\end{align}

We choose to parameterize $\textbf{r}_t | \textbf{x}_{t+1}, \textbf{x}_0$ and $\textbf{r}_t | \textbf{x}_{t+1}$ as vectors of iid Bernoulli random variables as this will allow us to analytically compute certain expectations in the ELBO, leading to a lower variance objective. This can be seen more clearly in the next section.

\subsection{Loss Function}
We train the model by maximising the ELBO, which for a single data point $\textbf{x}_0$ is defined as 

\begin{equation}
    \ln P_{\theta, \psi}(\textbf{x}_0) \geq \mathbb{E}_{Q_{\phi}(\mathbf{x}_{1:T}, \mathbf{r}_{0:T}|\mathbf{x}_0)} \left[ \log \left( \frac{P_{\theta, \psi}(\mathbf{x}_0, \mathbf{x}_{1:T}, \mathbf{r}_{0:T})}{Q_{\phi}(\mathbf{x}_{1:T}, \mathbf{r}_{0:T} | \mathbf{x}_0)} \right) \right] = \mathcal{L}(\mathbf{x}_0, \theta, \psi, \phi).
\end{equation}

Using the properties of conditional independence across timesteps and data dimensions, as well as the forms of the distributions in equations \ref{generative_model} and \ref{eq:q_VI_full}, we can write the ELBO as 
\begin{equation}
    \mathcal{L}(\mathbf{x}_0, \theta, \psi, \phi) = (T-1)\mathbb{E}_{\substack{Q_{\phi}(\textbf{x}_{\geq t+1} | \textbf{x}_0) \\ t \sim \text{Unif}(0, T-1)}}\left[L_t\right]
\end{equation}
where 
\begin{equation}\label{app_loss}
\begin{aligned}
    L_t &= \mathbb{E}_{Q_{\phi}(\mathbf{x}_{\geq t+1}, \mathbf{r}_{\geq t+1} | \mathbf{x}_0)} \left[\sum_{\substack{n=1 \\ \text{s.t. } x_{t+1}^{n}=m}}^{N}q_{\phi}^{t, n}(\mathbf{x}_{t+1}, \mathbf{x}_0) \log \mu_{\theta}^{n}(\mathbf{x}_{t+1})_{x_{0}^{n}} \right. \\
    &\qquad \qquad \qquad \qquad \qquad \left. \phantom{\sum_{n=1}^{N}} - D_{KL}\left(Q_{\phi}(r^n_t|\mathbf{x}_{t+1}, \mathbf{x}_0) || P_{\psi}(r^n_t|\mathbf{x}_{t+1})\right)\right].
\end{aligned}
\end{equation}

See appendix A for a derivation of this result. The loss function reveals the multiple purposes fulfilled by the posterior probabilities $q_{\phi}^{t, n}$. It determines the partially masked sequences that the denoiser and token position selector see during training, as well as weight the denoiser cross-entropy loss at each masked token position in proportion to its probability of being unmasked at the current timestep, and provide unmasking probabilities for the token selector to match.

In terms of the effect the denoiser and token selector have on $q_{\phi}^{t, n}$, we can see the first term in the expectation of equation (10) encourages $Q$ to learn unmasking orders which maximise the denoiser confidence in the ground truth tokens, by leveraging information about $x_0$. The KL-divergence term encourages $Q$ to maintain an unmasking schedule which can be replicated by the token selector $P_{\psi}$ used during inference, so that there is no mismatch between generation orders seen during training and inference.

Due to the presence of learned parameters $\psi$ in the distribution $Q$, we use REINFORCE to obtain an unbiased estimate of gradients of the loss. In order to reduce excessive variance of this estimator, we use REINFORCE-Leave-One-Out (RLOO) control variates \cite{kool2019buy}. Details of gradient estimation are in appendix B.

\subsection{Variational Posterior Design}
Up until this point, the form of the posterior unmasking probabilities $q^{t, n}$ has remained abstract. In this section, we describe the design choices explored in this work. First, we state some desirable properties that we want the posterior to satisfy:

\begin{enumerate}
    \item Sampling up to a given time $t \in \{0, ..., T-1\}$ should be computationally efficient (i.e., not scale significantly with $t$), so that we can easily compute unbiased Monte Carlo estimates of the loss at a randomly sampled time point during training,
    \item The posterior should be capable of parallel generation,
    \item The posterior should encode a notion of generation order i.e., that certain tokens should be generated before others.
    \item The posterior should unmask at least one token in each sampling step, in order to avoid wasted computation during training and inference
\end{enumerate}

To satisfy these properties, we compute the posterior unmasking probabilities $q_{\phi}^{t, n}(\textbf{x}_{t+1}, \textbf{x}_0)$ through a sequence of lightweight re-normalisation steps, based on an initial sequence of scores computed by a neural network $\alpha: \mathcal{V}^N \rightarrow \left[0, 1\right]^{N}$ with learnable paramaters $\phi$, 

\begin{equation}\label{eq:approx_post}
    q_{\phi}^{t, n}(\mathbf{x}_{t+1}, \mathbf{x}_0) = e^{\left(\alpha(\textbf{x}_0)_{n} - \text{Max}\{\alpha(\textbf{x}_0)_{n} | n \text{ s.t. } x_{t+1}^{n}=m\}\right)/\tau}.
\end{equation}

Note that this design choice satisfies property 1 as sampling requires a single pass through a neural network, followed by a sequence of $t$ updates of negligible cost. It satisfies property 2 as indices with similar $\alpha$ values have a high probability of being generated in the same step. Property 3 is satisfied by assigning higher $\alpha$ scores to tokens which should be unmasked earlier. Finally, property 4 is satisfied by the Max normalisation ensuring that at least one token is unmasked with probability 1 in each step. Empirically, we have found the inclusion of the temperature scaling parameter to be beneficial, typically setting it between 0.1 and 0.05 in our experiments. We hypothesise this may be due to one or both of the following reasons; 1) the temperature scaling prevents the randomly initialised $\alpha$ scores used at the start of training from unmasking in too few steps, leading to the majority of samples in a batch containing no training signal, 2) temperature scaling decreases the stochasticity in the unmasking orders generated by $Q$, leading to lower variance training and faster convergence.

\begin{table}[t]
\centering
\caption{\textbf{GSM8K Performance.} Comparison of our learned unmasking order against the baseline at equivalent average and maximum step costs. 
Since for $T=10$ our method's average number of steps rounds up to its maximum number of steps, we report the performance of baselines at the floor of this average.
Bold values in the final column represent maximum accuracy at the average number of steps that our method took after being trained with a given budget.}
\label{tab:budget_comparison}\begin{tabular}{@{}cllccc@{}}
\toprule
\textbf{Budget ($T$)} & \textbf{Method} & \textbf{Sampling Cost} & \textbf{Avg. Steps} & \textbf{Range} & \textbf{Acc. (\%)} \\ \midrule
\multirow{7}{*}{5} & IID & ID @ Avg. & 4.0 & [4, 4] & 29.0  \\
                    & IID & IID @ Max Used & 5.0 & [5, 5] & 29.9  \\
                    & Top Prob & Top Prob @ Avg. & 4.0 & [4, 4] & 23.7 \\
                    & Top Prob & Top Prob @ Max Used & 5.0 & [5, 5] & 26.6\\
                    & Top Prob Marg. & Top Prob Marg. @ Avg. & 4.0 & [4, 4] & 24.0  \\
                    & Top Prob Marg. & Top Prob Marg. @ Max Used & 5.0 & [5, 5] & 27.0  \\ \cmidrule(l){2-6} 
                    & \textbf{Ours} & \textbf{Learned Order} & \textbf{4.01} & \textbf{[2, 5]} & \textbf{33.1} \\ \midrule
\multirow{7}{*}{10} & IID & ID @ $\lfloor\text{Avg.}\rfloor$ & 9.0 & [9, 9] & 34.2 \\
                    & IID & IID @ Max Used & 10.0 & [10, 10] & 36.0  \\
                    & Top Prob & Top Prob @ $\lfloor\text{Avg.}\rfloor$ & 9.0 & [9, 9] & 35.9 \\
                    & Top Prob & Top Prob @ Max Used & 10.0 & [10, 10] & 37.8  \\
                    & Top Prob Marg. & Top Prob Marg. @ $\lfloor\text{Avg.}\rfloor$ & 9.0 & [9, 9] & 36.9  \\
                    & Top Prob Marg. & Top Prob Marg. @ Max Used & 10.0 & [10, 10] & \textbf{39.5}  \\ \cmidrule(l){2-6} 
                    & \textbf{Ours} & \textbf{Learned Order} & \textbf{9.57} & \textbf{[7, 10]} & 37.8 \\ \midrule
\multirow{7}{*}{15} & IID & ID @ Avg. & 9.0 & [9, 9] & 34.2  \\
                    & IID & IID @ Max Used & 12.0 & [12, 12] & 37.0  \\
                    & Top Prob & Top Prob @ Avg. & 9.0 & [9, 9] & 35.9 \\
                    & Top Prob & Top Prob @ Max Used & 12.0 & [12, 12] & 41.1 \\
                    & Top Prob Marg. & Top Prob Marg. @ Avg. & 9.0 & [9, 9] & 36.9  \\
                    & Top Prob Marg. & Top Prob Marg. @ Max Used & 12.0 & [12, 12] & 42.3  \\ \cmidrule(l){2-6} 
                    & \textbf{Ours} & \textbf{Learned Order} & \textbf{9.43} & \textbf{[5, 12]} & \textbf{39.0} \\ \midrule
\end{tabular}
\vfill
\end{table}

\section{Experiments}\label{sec:experiments}

We conduct experiments on the GSM8k dataset, initially following the same supervised finetuning method as Shin et. al. \cite{nieScalingMaskedDiffusion2025} on a 170M parameter MDM for 45,000 steps with batch size $256$, before carrying out further training with our algorithm using this pre-trained denoiser, and randomly initialised networks for $\alpha$ and $p_{\psi}^{t, n}$.
Specifically, we train according to the method laid out in Section \ref{sec:method} for an additional 15,000 steps using a batch size of 32 and drawing 8 RLOO generations per dataset sample. 
As a baseline for comparison, we continue finetuning the vanilla 170M MDM (without $\alpha$ and $p_{\psi}^{t,n}$ networks) on GSM8k, but with a batch size of 256 for 15,000 steps. 
This results in both our model and the baseline model seeing a batchsize of 256 samples, however, the baseline does get to see more diverse batches.  

To decode on the baseline model, we use a linear unmasking schedule to control the number of decoding steps and consider three standard sampling strategies, the first being the simplest approach from \cite{nieScalingMaskedDiffusion2025}, with the second and third suggested by \cite{kimTrainWorstPlan2025}. 
\begin{itemize}
    \item \textbf{IID}: Every masked token is unmasked independently with the same probability, determined by the linear unmasking schedule. 
    \item \textbf{Top Probability}: At each decoding step, the number of tokens to be unmasked, $K$, is determined by the linear unmasking schedule. Then we unmask at the $K$ indices for which our denoiser has maximal confidence about which token in the vocabulary, $\mathcal{V}$, to predict. That is, we unmask indices $i$ in $$\underset{i}{\text{Top-k}}(\max \mu_{\theta}^{i}(\textbf{x}_{t+1}, t)).$$
    \item \textbf{Top Probability Margin}: Similarly to \textbf{Top Probability}, we unmask $K$ masked tokens, but instead of choosing those with the highest confidence, we choose those with the largest difference between the probability assigned to the \textit{most likely} token, $j_1 \in \mathcal{V}$, and the \textit{second most likely} token, $j_2 \in \mathcal{V}$. That is, we unmask at indices $i$ in $$\underset{i}{\text{Top-k}}\left(\mu_{\theta}^{i}(\textbf{x}_{t+1}, t)_{j_1} - \mu_{\theta}^{i}(\textbf{x}_{t+1}, t)_{j_2}\right).$$
    This method attempts to improve upon \textbf{Top Probability} by not unmasking a token until there is a single most likely value for it to take, avoiding situations where multiple token-values have high-probability.
\end{itemize}


We compare the results of our method for multiple training budgets of $T$ (the maximum number of decoding steps). 
Since our method performs \textit{adaptive unmasking}, in the sense that it does not always use the same number of decoding steps for every prompt, we report the mean, min and max number of decoding steps on the GSM8k test set, and compare against the baseline methods evaluated at the mean and max values.

\section{Discussion}\label{sec:discussion} 
As can be seen in Table \ref{tab:budget_comparison}, our method successfully learns generation strategies that outperform the baseline generation methods at comparable or higher numbers of steps. 
This suggests that our method is indeed performing parallel generation in a manner that better avoids the pitfalls of over-parallelisation, especially when we consider the extremely low-budget T=5 setting.

One exception is the case where our model trained with a budget of $T=10$ ends up using an average of 9.57 steps and achieves slightly worse performance than top probability margin sampling with 10 steps (but better than the baselines with 9 steps).
Clearly the extra 0.43 steps give the top probability margin method a slight advantage, and so a direct comparison is difficult, but this does line up with the observation that the gap in performance between our method and the non-IID baselines closes as the decoding budget increases and the risk of over-parallelisation error decreases.

Whilst our results in Table \ref{tab:budget_comparison} demonstrate a  valid proof-of-concept for our method, a full analysis would require many more experiments in future work. 
For example, we would like to analyse the performance of our method on more datasets and with MDMs of varying sizes.
In the development of this paper, we experimented with a variety of approximate posterior forms before ending up with Eq.~\ref{eq:approx_post}, which seems to perform well.
Further experimentation of approximate posterior forms could be a fruitful direction for future work.

\section{Conclusion}
In this work, we present a method by which discrete diffusion models can decide the generation order of tokens at inference time, using a small learned auxiliary network. 
This allows the model to adaptively adjust its degree of parallelism based on the task at hand: whilst parallel generation is one of the main advantages of DDMs compared to standard ARMs, generation schemes which lead to too much parallelism can hurt downstream task performance \citep{kang2025parallelbench}.
We explore a strategy for learning this auxiliary network via variational inference, wherein we treat the generation order as a latent variable to be inferred, and show that this method achieves competitive results against standard baseline MDM generation schemes on GSM8k.
We anticipate that further analysis and development of this method would be very helpful towards improving the performance of discrete diffusion models at large.



\bibliography{bibliography}
\bibliographystyle{bib_style}

\appendix
\section{Appendix}

\subsection{Masked Diffusion Loss Function Derivation}
The generative model is trained using the ELBO of $P_{\theta, \psi}$, using amortised posterior $Q_{\phi}$,

\begin{equation}
    \mathcal{L}(\mathbf{x}_0, \theta, \psi, \phi) = \mathbb{E}_{Q_{\phi}(\mathbf{x}_{1:T}, \mathbf{r}_{0:T}|\mathbf{x}_0)} \left[ \log \left( \frac{P_{\theta, \psi}(\mathbf{x}_0, \mathbf{x}_{1:T}, \mathbf{r}_{0:T})}{Q_{\phi}(\mathbf{x}_{1:T}, \mathbf{r}_{0:T} | \mathbf{x}_0)} \right) \right].
\end{equation}

In what follows we use the notation $\mathbf{x}_{\geq t} = \mathbf{x}_{t:T-1}$. 

Applying conditional independence over timesteps $t$ gives,

\begin{align*}
    \mathcal{L}(\mathbf{x}_0, \theta, \psi, \phi) &= \mathbb{E}_{Q_{\phi}(\mathbf{x}_{\geq1}, \mathbf{r}_{\geq0:}|\mathbf{x}_0)}\left[\log P_{\theta, \psi}(\mathbf{x}_0, \mathbf{r}_0 | \mathbf{x}_1)\right] + \mathbb{E}_{Q_{\phi}(\mathbf{x}_{\geq 1}, \mathbf{r}_{\geq 0}|\mathbf{x}_0)}\left[\log \prod_{t=1}^{T-1} \frac{P_{\theta, \psi}(\mathbf{x}_t, \mathbf{r}_t | \mathbf{x}_{t+1})}{Q_{\phi}(\mathbf{x}_t, \mathbf{r}_t | \mathbf{x}_{t+1}, \mathbf{x}_0)}\right] \\
    &= \mathbb{E}_{Q_{\phi}(\mathbf{x}_{\geq 1}, \mathbf{r}_{\geq 0}|x_0)}\left[\log P_{\theta, \psi}(\mathbf{x}_0, \mathbf{r}_0 | \mathbf{x}_1)\right]+ \sum_{t=1}^{T-1} \mathbb{E}_{Q_{\phi}(\mathbf{x}_{\geq t}, \mathbf{r}_{\geq t}|\mathbf{x}_0)} \left[ \log\frac{P_{\theta, \psi}(\mathbf{x}_t, \mathbf{r}_t | \mathbf{x}_{t+1})}{Q_{\phi}(\mathbf{x}_t, \mathbf{r}_t | \mathbf{x}_{t+1}, \mathbf{x}_0)} \right] \\
    &= L_{0} + \sum_{t=1}^{T-1}L_t.
\end{align*}

Applying the factorization over $\mathbf{x}_t, \mathbf{r}_t$ in $P$ and $Q$ to the $L_t$ terms gives

\begin{align*}
    L_t &= \mathbb{E}_{Q_{\phi}(\mathbf{x}_{\geq t+1}, \mathbf{r}_{\geq t + 1}|\mathbf{x}_0)} \left[ \mathbb{E}_{Q(\mathbf{x}_t|\mathbf{r}_t, \mathbf{x}_{t+1}, \mathbf{x}_0)Q_{\phi}(\mathbf{r}_t|\mathbf{x}_{t+1}, \mathbf{x}_0)} \left[ \log \frac{P_{\theta}(\mathbf{x}_t|\mathbf{r}_t, \mathbf{x}_{t+1})}{Q(\mathbf{x}_t|\mathbf{r}_t, \mathbf{x}_{t+1}, \mathbf{x}_0)} \right] \right. \\
    &\qquad \left. + \mathbb{E}_{Q(\mathbf{x}_t|\mathbf{r}_t, \mathbf{x}_{t+1}, \mathbf{x}_0)Q_{\phi}(\mathbf{r}_t|\mathbf{x}_{t+1}, \mathbf{x}_0)} \left[ \log \frac{P_{\psi}(\mathbf{r}_t|\mathbf{x}_{t+1})}{Q_{\phi}(\mathbf{r}_t|\mathbf{x}_{t+1}, \mathbf{x}_0)} \right] \right] \\
    & = \mathbb{E}_{Q_{\phi}(\mathbf{x}_{\geq t+1}, \mathbf{r}_{\geq t + 1}|\mathbf{x}_0)} \left[ - \mathbb{E}_{Q_{\phi}(\mathbf{r}_t | \mathbf{x}_{t+1}, \mathbf{x}_0)} \left[ D_{KL}\left(Q(\mathbf{x}_t|\mathbf{r}_t, \mathbf{x}_{t+1}, \mathbf{x}_0) || P_{\theta}(\mathbf{x}_t|\mathbf{r}_t, \mathbf{x}_{t+1})\right) \right] \right. \\
    &\qquad \left. - D_{KL}\left(Q_{\phi}(\mathbf{r}_t|\mathbf{x}_{t+1}, \mathbf{x}_0) || P_{\psi}(\mathbf{r}_t|\mathbf{x}_{t+1})\right) \right].
\end{align*}

Applying conditional independence over dimensions of $\textbf{x}_t$ given $\textbf{r}_t$ gives,

\begin{align*}
    L_t &= \mathbb{E}_{Q_{\phi}(\mathbf{x}_{\geq t+1}, \mathbf{r}_{\geq t+1}|\mathbf{x}_0)} \left[ -\mathbb{E}_{Q_{\phi}(\mathbf{r}_t | \mathbf{x}_{t+1}, \mathbf{x}_0)} \left[ \sum_{n=1}^{N} D_{KL}\left(Q(x_t^{n}|r_t^{n}, x_{t+1}^{n}, x_0^{n}) || P_{\theta}(x_t^{n}|r_t^{n}, x_{t+1}^{n})\right) \right] \right. \\
    &\qquad \left. - D_{KL}\left(Q_{\phi}(\mathbf{r}_t|\mathbf{x}_{t+1}, \mathbf{x}_0) || P_{\psi}(\mathbf{r}_t|\mathbf{x}_{t+1})\right) \right].
\end{align*}

For $Q$ and $P_{\theta}$ given by 

\begin{align*}
    & Q(x_t^{n} | r_t^{n}, x_{t+1}^{n}, x_0^{n}) = \delta_{x_{t+1}^{n}}(1 - r_t^{n}) + \delta_{x_0^{n}} {r_t^{n}}, \\
    & P_{\theta}(x_t^{n} | r_t^{n}, \mathbf{x}_{t+1}) = \delta_{x_{t+1}^{n}}(1 - r_t^{n}) + \mu_{\theta}^{n}(\textbf{x}_{t+1}, t) r_t^{n},
\end{align*}

the KL divergences are given in closed form by 

\begin{align*}
D_{KL}\left(Q(x_t^{n} | r_t^{n}, x_{t+1}^{n}, x_0^{n}) \ || \ P_{\theta}(x_t^{n} | r_t^{n}, \mathbf{x}_{t+1})\right) &= D_{KL}\left(\delta_{x_0^{n}} \ || \ \mu_{\theta}^{n}(\mathbf{x}_{t+1}, t)\right) r_t^{n}, \\
&=  -r_t^{n} \log \mu_{\theta}^{n}(x_{t+1}, t)_{x_{0}^{n}},
\end{align*}

subbing this into $L_t$ gives

\begin{align}\label{app_loss}
        L_t &= \mathbb{E}_{Q_{\phi}(\mathbf{x}_{\geq t+1}, \mathbf{r}_{\geq t+1}|\mathbf{x}_0)} \left[ \mathbb{E}_{Q_{\phi}(\mathbf{r}_t | \mathbf{x}_{t+1}, \mathbf{x}_0)} \left[ \sum_{n=1}^{N} r_t^{n} \log \mu_{\theta}^{n}(\textbf{x}_{t+1}, t)_{x_{0}^{n}} \right] \right. \\
    &\qquad \left. - D_{KL}\left(Q_{\phi}(\mathbf{r}_t|\mathbf{x}_{t+1}, \mathbf{x}_0) || P_{\psi}(\mathbf{r}_t|\mathbf{x}_{t+1})\right) \right]
\end{align}

We assume conditionally independent $r^{n}_t$, as this leads to a lower variance final expression for $L_t$, where expectations with respect to all $r^{n}_t$ are computed in closed form. The conditional distributions for $\textbf{r}_t$ are therefore paramaterised as follows,

\begin{align*}
    Q_{\phi}(\mathbf{r}_t | \mathbf{x}_{t+1}, \mathbf{x}_0) &= \prod_{n=1}^{N} Q_{\phi}(r_t^{n} | \mathbf{x}_{t+1}, \mathbf{x}_0), \\
    P_{\theta}(\mathbf{r}_t | \mathbf{x}_{t+1}) &= \prod_{n=1}^{N} P_{\theta}(r_t^{n} | \mathbf{x}_{t+1}),
\end{align*}

where

\begin{align*}
    \quad Q_{\phi}(r_t^{n} | \mathbf{x}_{t+1}, \mathbf{x}_0) &= \text{Bern}\left(q_{\phi}^{t, n}(\mathbf{x}_{t+1}, \mathbf{x}_0) \textbf{1}_{x_{t+1}^{n}=m}\right), \\
    \quad P_{\theta}(r_t^{n} | \mathbf{x}_{t+1}) &= \text{Bern}\left(p_{\psi}^{t, n}(\mathbf{x}_{t+1}) \textbf{1}_{x_{t+1}^{n}=m}\right).
\end{align*}

The probabilities $p_{\psi}^{t, n}$ and $q_{\phi}^{t, n}$ are implemented using neural networks. We can now simplify each term in the outer expectation of \eqref{app_loss},

\begin{align*}
    \mathbb{E}_{Q_{\phi}(\mathbf{r}_t | \mathbf{x}_{t+1}, \mathbf{x}_0)} \left[ \sum_{n=1}^{N} r_t^{n} \log \mu_{\theta}^{n}(\mathbf{x}_{t+1}, t)_{x_{0}^{n}} \right] &= \sum_{n=1}^{N}\mathbb{E}_{\prod_{n=1}^{N} Q_{\phi}(r_t^{n} | \mathbf{x}_{t+1}, \mathbf{x}_0)} \left[ r_t^{n} \log \mu_{\theta}^{n}(\mathbf{x}_{t+1}, t)_{x_{0}^{n}} \right] \\
    &= \sum_{\substack{n=1, \\\text{s.t.} x_{t+1}^{n}=m}}^{N}q^{t, n}(\mathbf{x}_{t+1}, \mathbf{x}_0) \log \mu_{\theta}^{n}(\mathbf{x}_{t+1})_{x_{0}^{n}}.
\end{align*}

\begin{align*}
D_{KL}\left(Q_{\phi}(\mathbf{r}_t | \mathbf{x}_{t+1}, \mathbf{x}_0) \ || \ P_{\theta}(\mathbf{r}_t | \mathbf{x}_{t+1})\right) &= D_{KL}\left(\prod_{n=1}^{N} Q_{\phi}(r_t^{n} | \mathbf{x}_{t+1}, \mathbf{x}_0) \ || \ \prod_{n=1}^{N} P_{\theta}(r_t^{n} | \mathbf{x}_{t+1})\right), \\
&= \sum_{\substack{n=1, \\\text{s.t.} x_{t+1}^{n}=m}}^{N} D_{KL}\left(Q_{\phi}(r_t^{n} | \mathbf{x}_{t+1}, \mathbf{x}_0) \ || \ P_{\theta}(r_t^{n} | \mathbf{x}_{t+1})\right) \\
\end{align*}

Where in the last line we've used the fact that the KL divergences in the sum are 0 if $x_t^{n} \neq m$.

Subbing these expressions into equation (13) then gives, 

\begin{align}\label{app_loss}
        L_t &= \mathbb{E}_{Q(\mathbf{x}_{\geq t+1}, \mathbf{r}_{\geq t+1} | \mathbf{x}_0)} \left[\sum_{\substack{n=1, \\\text{s.t.} x_{t+1}^{n}=m}}^{N}q^{t, n}(\mathbf{x}_{t+1}, \mathbf{x}_0) \log \mu_{\theta}^{n}(\mathbf{x}_{t+1})_{x_{0}^{n}} \right. \\
        &\qquad \qquad \qquad \qquad \qquad \qquad \left. - D_{KL}\left(Q(r^n_t|\mathbf{x}_{t+1}, \mathbf{x}_0) || P_{\psi}(r^n_t|\mathbf{x}_{t+1})\right)\right].
\end{align}

The finial total loss function is then

\section{Masked Diffusion Gradient Estimation}
We first introduce some notations for the sake of brevity,

\begin{align*}
    F^{1}(\mathbf{x}_{t+1}, \textbf{x}_0,\textbf{q}_{\phi}^{t}) &= \sum_{\substack{n=1 \\ \text{s.t. } x_{t+1}^{n}=m}}^{N}q_{\phi}^{t, n}(\mathbf{x}_{t+1}, \mathbf{x}_0) \log \mu_{\theta}^{n}(\mathbf{x}_{t+1})_{x_{0}^{n}}, \\
    F^{2}(\mathbf{x}_{t+1}, \mathbf{q}_{\phi}^{t}, \mathbf{p}_{\psi}^{t}) &= - \sum_{\substack{n=1 \\ \text{s.t. } x_{t+1}^{n}=m}}^{N}D_{KL}\left(Q_{\phi}(r^n_t|\mathbf{x}_{t+1}, \mathbf{x}_0) || P_{\psi}(r^n_t|\mathbf{x}_{t+1})\right), \\
    F(\mathbf{x}_{t+1}, \mathbf{x}_0, \mathbf{q}_{\phi}^{t}, \mathbf{p}_{\psi}^{t}) &= F^{1}(\mathbf{x}_{t+1}, \textbf{x}_0,\textbf{q}_{\phi}^{t}) +  F^{2}(\mathbf{x}_{t+1}, \mathbf{q}_{\phi}^{t}, \mathbf{p}_{\psi}^{t}),
\end{align*}

where we suppress dependence of $p_{\phi}^{t, n}$ and $p_{\psi}^{t, n}$ on $\textbf{x}_{t+1}$ and $\textbf{x}_0$ for notational simplicity. We additionally denote the vector of unmasking probabilities over token indices using bold typeface. 

The gradients of the loss are computed as follows,

\begin{align*}
    \eta_{\psi}(\mathbf{x}_0, \psi, \phi) &= \nabla_{\psi}\mathcal{L}(\mathbf{x}_0, \psi, \phi), \\
    &=\sum_{t=1}^{T-1} \mathbb{E}_{Q_{\phi}(\mathbf{x}_{\geq t+1} | \textbf{x}_0)} \left[ \nabla_{\psi}F^{2}(\mathbf{x}_{t+1}, \mathbf{q}_{\phi}^{t}, \mathbf{p}_{\psi}^{t}) \right], \\
    &= (T-1)\mathbb{E}_{\substack{Q_{\phi}(\mathbf{x}_{\geq t+1} | \mathbf{x}_0) \\ t \sim \text{Unif}(1, T-1)}} \left[ \nabla_{\psi}F^{2}(\mathbf{x}_{t+1}, \mathbf{q}_{\phi}^{t}, \mathbf{p}_{\psi}^{t}) \right],
\end{align*}

\begin{align*}
    \eta_{\phi}(\mathbf{x}_0, \psi, \phi) &= \nabla_{\phi}\mathcal{L}(x_0, \psi, \phi), \\
    &= (T-1)\mathbb{E}_{\substack{Q_{\phi}(\textbf{x}_{\geq t+1} | \textbf{x}_0) \\ t \sim \text{Unif}(1, T-1)}} \left[ \nabla_{\phi}F(\mathbf{x}_{t+1}, \mathbf{x}_0, \mathbf{q}_{\phi}^{t}, \mathbf{p}_{\psi}^{t}) \right], \\
    &\qquad + (T-1)\mathbb{E}_{\substack{Q_{\phi}(\mathbf{x}_{\geq t+1} | \mathbf{x}_0) \\ t \sim \text{Unif}(1, T-1)}} \left[F(\mathbf{x}_{t+1}, \mathbf{x}_0, \mathbf{q}_{\phi}^{t}, \mathbf{p}_{\psi}^{t}) \nabla_{\phi}{\log Q_{\phi}(\mathbf{x}_{\geq t+1} | \textbf{x}_0)}\right], \\
    &= \eta_{\phi}^{1}(\mathbf{x}_0, \psi, \phi) + \eta^{2}_{\phi}(\mathbf{x}_0, \psi, \phi) \\
\end{align*}

where we've used the log-derivative trick / REINFORCE to compute $\eta_{\phi}$. $\eta_{\psi}$ and $\eta_{\phi}^{1}$ are expectations of gradients, and can be estimated using naive monte carlo. 

\begin{align*}
     \bar{\eta}^{k}_{\psi}(\mathbf{x}_0, \psi, \phi) &= \frac{T-1}{k}\sum_{i=1}^{k}\nabla_{\psi}F^{2}(\hat{\mathbf{x}}^{i}_{t+1}, \mathbf{q}_{\phi}^{t}, \mathbf{p}_{\psi}^{t}), \\
     t &\sim \text{Unif}(1, \dots, T-1), \\
\left(\hat{\textbf{x}}^{i}_{\geq t+1}, \hat{\textbf{r}}^{i}_{\geq t+1}\right) &\sim Q_{\phi}(\mathbf{x}_{\geq t+1}, \mathbf{r}_{\geq t+1} | \mathbf{x}_0), \quad i = 1, ..., k.
\end{align*}

In $\eta_{\phi}^{2}$, the gradient in the expectation is scaled by the sum of $F$'s, which typically leads to excessively high variance in a naive monte carlo estimate. Overcoming this requires variance reduction techniques. For simplicity, we use REINFORCE Leave-One-Out (RLOO) \cite{kool2019buy}. 

\begin{align*}
\bar{\eta}_{\phi}^{2, k} &= \frac{T-1}{k} \sum_{i=1}^{k} \left[ F(\hat{\mathbf{x}}^{i}_{t+1}, \mathbf{x}_0, \mathbf{q}_{\phi}^{t}, \mathbf{p}_{\psi}^{t}) \right. \\
& \qquad \left. - \frac{1}{k-1} \sum_{l=1,\substack{ l \neq j=i}}^{k} F(\hat{\mathbf{x}}^{l}_{t+1}, \mathbf{x}_0, \mathbf{q}_{\phi}^{t}, \mathbf{p}_{\psi}^{t})\right] \nabla_{\phi}{\log Q_{\phi}(\hat{\mathbf{x}}^{i}_{\geq t+1} | \textbf{x}_0)} \\
     t &\sim \text{Unif}(1, \dots, T-1), \\
\left(\hat{\textbf{x}}^{i}_{\geq t+1}, \hat{\textbf{r}}^{i}_{\geq t+1}\right) &\sim Q_{\phi}(\mathbf{x}_{\geq t+1}, \mathbf{r}_{\geq t+1} | \mathbf{x}_0), \quad i = 1, ..., k.
\end{align*}

\end{document}